\documentclass[10pt,conference]{IEEEtran}
\IEEEoverridecommandlockouts
\usepackage{cite}
\usepackage{amsmath,amssymb,amsfonts}
\usepackage{algorithmic}
\usepackage{graphicx}
\usepackage{textcomp}
\usepackage{xcolor}
\usepackage{multirow}
\usepackage[left=1.62cm,right=1.62cm,top=1.9cm]{geometry}

\def\BibTeX{{\rm B\kern-.05em{\sc i\kern-.025em b}\kern-.08em
    T\kern-.1667em\lower.7ex\hbox{E}\kern-.125emX}}
 \def\BibTeX{{\rm B\kern-.05em{\sc i\kern-.025em b}\kern-.08em T\kern-.1667em\lower.7ex\hbox{E}\kern-.125emX}}
\begin{document}

\title{A Two-Stage Multimodal Emotion Recognition Model Based on Graph Contrastive Learning\\
}

\author{
	\IEEEauthorblockN{Wei Ai}
	\IEEEauthorblockA{\textit{School of Computer and Information Engineering} \\
		\textit{Central South University of Forestry and Technology}\\
		ChangSha, China \\
		aiwei@hnu.edu.cn}
	\and
	\IEEEauthorblockN{FuChen Zhang}
	\IEEEauthorblockA{\textit{School of Computer and Information Engineering} \\
		\textit{Central South University of Forestry and Technology}\\
		ChangSha, China \\
		fuchen.zhang@csuft.edu.cn}
	\and
	\IEEEauthorblockN{\hspace{5em}Tao Meng{*}}
		\IEEEauthorblockA{\textit{\hspace{4em}School of Computer and Information Engineering}\\
		\textit{\hspace{4em}Central South University of Forestry and Technology}\\
		\hspace{4em}ChangSha, China \\
		\hspace{4em}mengtao@hnu.edu.cn}
	\and
	\IEEEauthorblockN{YunTao Shou}
	\IEEEauthorblockA{\textit{School of Computer and Information Engineering} \\
		\textit{Central South University of Forestry and Technology}\\
		ChangSha, China \\
		yuntaoshou@csuft.edu.cn}
	\and
    \IEEEauthorblockN{\hspace{4em}HongEn Shao}
	\IEEEauthorblockA{\textit{\hspace{4em}School of Computer and Information Engineering} \\
		\textit{\hspace{4em}Central South University of Forestry and Technology}\\
		\hspace{4em}ChangSha, China \\
		\hspace{4em}hongen.shao@csuft.edu.cn}
	\and
	\IEEEauthorblockN{\hspace{2em}Keqin Li}
	\IEEEauthorblockA{\hspace{3em}Department of Computer Science \\
		\hspace{3em}State University of New York\\
		\hspace{3em}New Paltz, New York 12561, USA \\
		\hspace{4em}lik@newpaltz.edu}
	\thanks{* is the corresponding author.}
}

\maketitle

\begin{abstract}
In terms of human-computer interaction, it is becoming more and more important to correctly understand the user's emotional state in a conversation, so the task of multimodal emotion recognition (MER) started to receive more attention. However, existing emotion classification methods usually perform classification only once. Sentences are likely to be misclassified in a single round of classification. Previous work usually ignores the similarities and differences between different morphological features in the fusion process. To address the above issues, we propose a two-stage emotion recognition model based on graph contrastive learning (TS-GCL). First, we encode the original dataset with different preprocessing modalities. Second, a graph contrastive learning (GCL) strategy is introduced for these three modal data with other structures to learn similarities and differences within and between modalities. Finally, we use MLP twice to achieve the final emotion classification. This staged classification method can help the model to better focus on different levels of emotional information, thereby improving the performance of the model. Extensive experiments show that TS-GCL has superior performance on IEMOCAP and MELD datasets compared with previous methods.
\end{abstract}


\section{Introduction}
\begin{figure}[htbp]
	\centering
	\includegraphics[width=0.94\linewidth,scale=1.00]{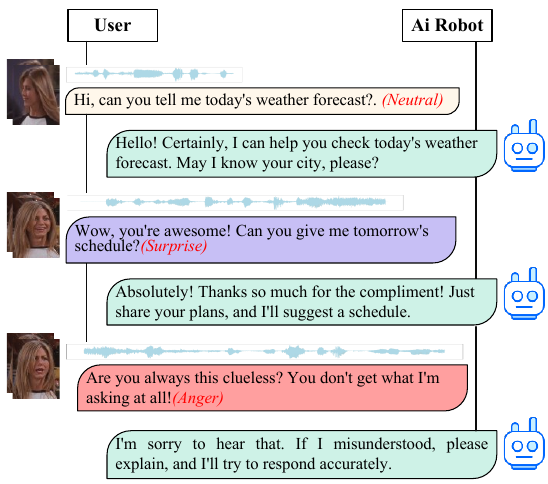}
	\caption{An example of effective multimodal multi-emotion human-machine interaction in which multimodal emotion recognition plays a key role.}
	\label{fig_1}
\end{figure}
Emotion is an indispensable element in human daily communication, and the goal of multimodal emotion recognition is to automatically identify and track the emotional state of speakers in a conversation \cite{17}. This task is receiving increasing attention in the fields of natural language processing (NLP) and multimodal processing. Multimodal emotion recognition has many potential applications, such as assisting dialogue analysis in legal trials and e-health services. Additionally, it is a critical component in creating more natural human-machine interactions, a specific example of which is shown in Fig. 1.

With the rapid growth of conversational data on social media platforms, more and more researchers have begun to pay attention to multimodal emotion recognition. Multimodal emotion recognition analyzes complex emotional expressions more easily. By combining multiple modalities, the diversity and complexity of human emotions can be better captured.

However, multimodal data such as video, audio, and text have differences in feature space distribution, which leads to the gap problem between different modalities in multimodal emotion recognition \cite{19}. To eliminate the differences between the features of each modality, the current mainstream multi-modal feature fusion method is to directly map each modality into a shared feature space for representation. For example, the Tensor Fusion Network (TFN) \cite{22} introduces tensor decomposition technology to decompose the fused multimodal feature representation into different weight matrices to capture the relationship between multimodal features, which helps to capture multimodal data interaction information better.  Nevertheless, the above methods still suffer from indelible heterogeneity when mapping different modal features to a standard representation space. To address the above issues, we introduce a strategy of graph contrastive learning to eliminate this heterogeneity.

In addition, the current deep learning methods often perform one-time classification when dealing with emotion recognition tasks, and there is no secondary correction or reclassification mechanism, which directly reduces the model's performance. For example, \textit{Hu et al.} \cite{24} used Graph Convolution Network (GCN) to achieve multimodal data fusion and emotion recognition. Two-stage classification can decompose the task into two smaller subtasks, thereby reducing the complexity of the overall mission and making the model easier to train and optimize. 

To address the above challenges, we propose a novel modal called a two-stage emotion recognition in conversations model based on graph contrastive learning (TS-GCL), which aims to simulate the emotion recognition process in human dialogue. Human beings can recognize the emotion of the sentence itself and modify or re-understand the feeling of a sentence according to the context. This process can be better simulated by using two classifications.

The main contributions of our study are summarized below:
\begin{itemize}
	\item[$\bullet$] We propose a novel two-stage multimodal emotion recognition model (TS-GCL) based on graph contrastive learning, which utilizes graph contrastive learning strategies to continuously update and correct the differences between samples, making the model anti-noise and anti-bias ability. At the same time, it effectively enhances the robustness of the model.
	
	\item[$\bullet$] We propose a novel two-stage classification method for MER, which has a clear division of labor. The first stage is to judge the emotional polarity (positive, negative, or neutral), and the second stage is to classify more detailed dynamic categories. Each stages is responsible for tasks of different granularity. The two-stage classification method is closer to human beings identifying emotions.

	\item[$\bullet$] Our proposed model is extensively evaluated on two benchmark datasets, IEMOCAP and MELD. The experimental results show its superiority over existing algorithms in accuracy and F1-score.

\end{itemize}

\section{Related work}

\subsection{Multimodal Emotion Recognition}

Emotion recognition has been fully applied in many fields, including social media analysis, customer service and market research, and public opinion monitoring.  With the development of artificial intelligence, deep learning methods have surpassed traditional machine learning algorithms in all aspects.

In recent years, as an essential topic of natural language processing, MER has attracted more and more interest from researchers \cite{33, 34, 35, 36, 37, 38, 39, 40, 41}. Unlike ordinary emotion recognition, MER needs to consider modeling the speaker's contextual information and the problem of semantic information fusion from multiple modalities. To this end, \textit{Poria et al.} \cite{12} proposed a video emotion recognition method that considers the context of video content. Specifically, they used LSTM for context modeling, and the context information was further used for emotion recognition. \textit{Hazarika et al.} \cite{13} proposed a Conversational Memory Network (CMM) to capture context dependencies in conversations. Although previous methods have achieved good performance on MER, most ignore the differences between different emotion categories, so we introduce a graph contrastive learning mechanism and a two-stage classification method, which can not only effectively model the emotion from three modal graph structure information. We can better learn the representation of graphs and compare graphs so that the features between graphs expressing the same emotion are more similar. The graph features of different emotions are more differentiated.
\subsection{Contrastive Learning}

Self-supervised learning (SL), as an essential part of deep learning, has received extensive attention in recent years. Contrastive representation learning (CRL) is a representative method in self-supervised learning. Its core idea is to learn the discriminative features for distinguishing samples by continuously reducing the distance between positive samples and expanding the space between positive samples.

\textit{Li et al.} \cite{14} proposed introducing contrastive representation learning into the model, randomly sampling multiple slices on the feature sequence, maximizing the similarity between different slice representations of the same speech, and minimizing other Similarity between slice representations. \textit{Kim et al.} \cite{15} proposed a Contrastive Adversarial Learning (CAL) framework, which consists of a Contrastive Learning Module and an Adversarial Module, to learn representations that distinguish between different expressions. Compared with methods that directly use contrastive learning for expression recognition, contrastive adversarial learning improves the robustness of features. Although the model's performance is further improved after introducing contrastive learning in previous methods, how to model the dependencies and internal consistency among different modalities becomes a new challenge.
\begin{figure*}[htbp]
	\centering
	\includegraphics[width=1.0\linewidth]{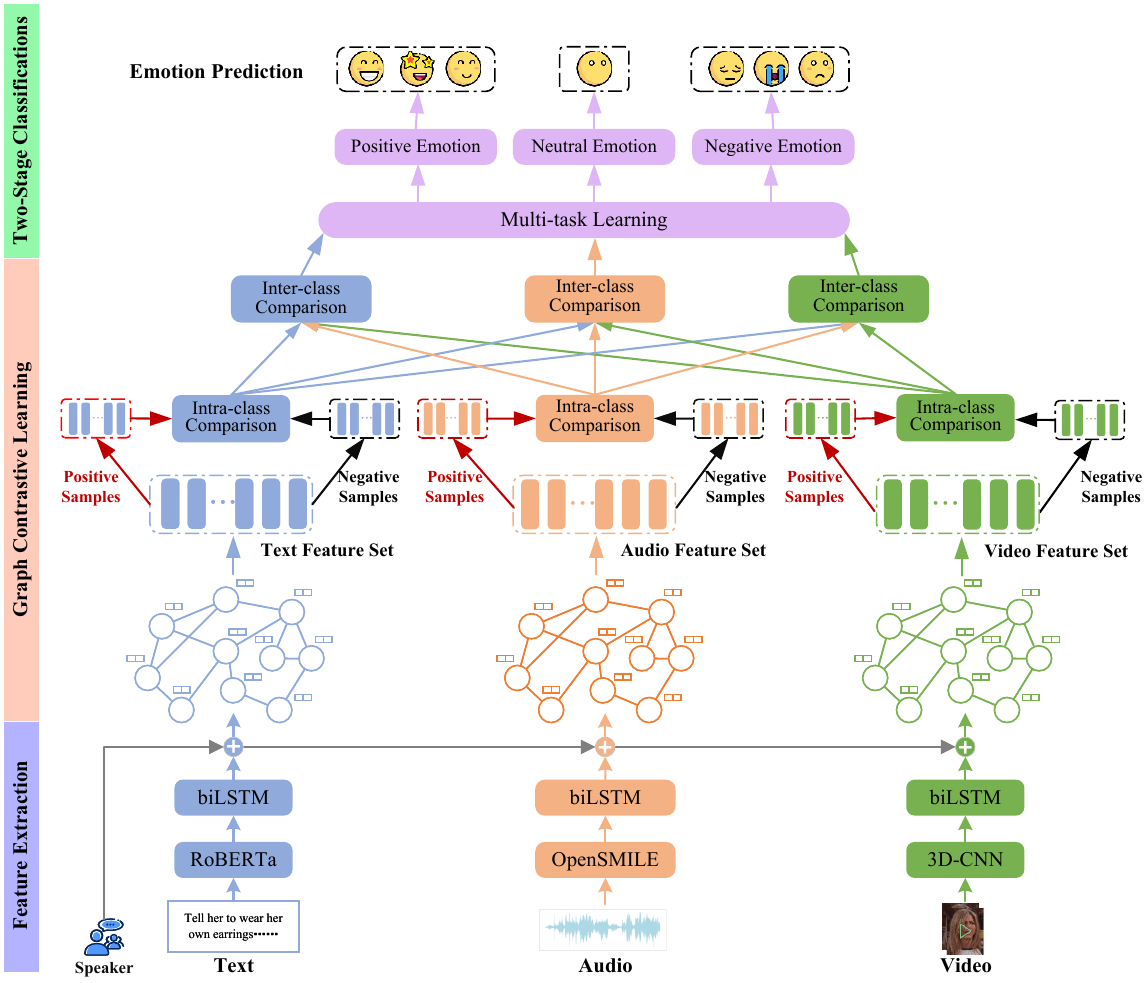}
	\caption{We propose the architecture of TS-GCL. It is mainly divided into three parts. The first part is feature extraction, using different preprocessing methods to process the original dataset. The second part is graph contrastive learning. This part describes in detail the graph construction process and the process of contrastive learning. The last part is two-stage classification, in which MLP is used for secondary classification in the emotion classification process, so as to achieve better classification results.}
	\label{fig_2}
\end{figure*}

\section{Preliminary Information}

In this section, we detail the use of different preprocessing methods for other modalities, and at the same time, we define the MER task mathematically.

\subsection{Feature Extraction}
To verify the model's performance, this paper uses two multimodal emotion recognition general data sets, IEMOCAP and MELD. They use three data forms (text, audio, video) for storage. Aiming at these three different data storage forms, a targeted data preprocessing method is used for feature encoding to obtain high-quality semantic feature representations with rich semantic information. We will describe the encoding process of each mode in detail as follows.
\subsubsection{Text Feature Extraction}

We will extract word-level text vector features from transcripts in the dataset. Specifically, influenced by previous work \cite{1}, \cite{2},\cite{33}, we convert transcripts into text vector features using the RoBERTa \cite{4} pre-trained model, which is the most advanced sentence encoding model known to us and has shown its superiority in a variety of tasks. Properties, including question-answering systems, named entity recognition, and information retrieval. Using this model, we obtain excellent word embedding representations $\delta^t$.

\subsubsection{Audio Feature Extraction}

The fluctuation of sound in the audio signal is not only the expression of sound but also reflects the ups and downs of the speaker's inner emotions. Sometimes, a person's behavior may not accurately reflect his emotional state, but his voice is an undisguised expression of real emotion. Inspired by previous research \cite{5}, \cite{6}, we extracted audio features $\delta^a$ using OpenSMILE.

\subsubsection{Vision Feature Extraction}

We utilize 3D-CNN as a visual feature extractor. The model is a network architecture for visual representation learning. Recent work \cite{5}, \cite{6} has demonstrated its capability in various downstream tasks such as video analysis and medical image processing. Finally, we used 3D-CNN to extract 512-dimensional visual features $\delta^v$.

\subsection{Problem Definition}

Suppose $I$ speakers and $N$ sentences are in a multimodal dialogue. Then it can be expressed as two sets $S$ and $U$, where $S=\{s_1,s_2,\ldots,s_I\}$, $U=\{u_1,u_2,\ldots,u_N\}$. The multimodal emotion recognition task needs to predict the emotional label of each sentence when the speaker speaks. We define a mapping function $\varphi$ to represent the connection between each sentence and the speaker. For example, $s_{\varphi_{n}}$ represents the speaker corresponding to the $n$-th sentence. We will use the preprocessing method described above to obtain the representation features $u_i\in\mathbb{R}^d$ of the utterance. The representation features of each utterance contain data from three modalities, namely text, audio, and vision. In summary, it can be expressed as:
\begin{equation}
	u_i=\{\delta_i^t,\delta_i^a,\delta_i^v\}
\end{equation}

\section{Proposed method}

In order to enhance the performance of multimodal emotion recognition, we propose a new method called a two-stage multimodal emotion recognition model based on graph contrastive learning (TS-GCL). The overall architecture of TS-GCL is shown in Fig. 2.

\subsection{Modality Encoder}

Contextual information plays an important role in predicting emotion labels for each utterance. Therefore, converting contextual information into discourse feature expression has positive benefits. We use a bidirectional Long Short Term Memory (LSTM) for contextual information extraction on sentence sequences $U=\{u_1,u_2,\ldots,u_N\}$ because it can effectively capture long-term dependencies, which is important for understanding long-distance associations in text or sequence data. The calculation process is as follows:

\begin{equation}
	\overrightarrow{h_i} =\overrightarrow{LSTM}(u_i),
\end{equation}
\begin{equation}
	\overleftarrow{h_i}=\overleftarrow{LSTM}(u_i),
\end{equation}
\begin{equation}
	h_i=[\overrightarrow{h_i},	\overleftarrow{h_i}],i\in\{1,\dots,N\}.
\end{equation}
where $\overrightarrow{h_i}$ and $\overleftarrow{h_i}$ represent the cell states of two unidirectional lstm forward propagation and backward propagation respectively, $h_{i}\in\mathbb{R}^{d}$ represents the output of a bidirectional LSTM.

\subsection{Speaker Embedding}

Speaker information can provide clues about the speaker's social background, emotional tendencies, and attitudes. This helps to better understand the context of the text and thus predict emotion more accurately. And speaker information can enhance the modeling of dialogue context. The same piece of text may have different emotional interpretations between different speakers. Embedding speaker information helps to capture context information more accurately. Therefore, we decide to embed the original speaker's information $s_i$ into it before composing it. The speaker embedding $\lambda_i$ calculation process is as follows:

\begin{equation}
	\lambda_i=W_s s_i+b_i^\lambda
\end{equation}
where $W_s$ is the weight matrix and $b_i^\lambda$ is the bias vector.
\subsection{Graph Construction}

Inspired by previous work \cite{31},\cite{32},\cite{30}. we construct a directed graph $\mathcal{G}_{\mathcal{M}}=\{\mathcal{V}_{\mathcal{M}},\mathcal{E}_{\mathcal{M}},\mathcal{T}_{\mathcal{M}}\}$ to represent a dialogue consisting of $I$ utterances as follows, where $\mathcal{M}\in\{t,a,v\}$. In this graph, the node set $\mathcal{V}(|\mathcal{V}|=3N)$ corresponds to utterances in three different utterance modes, and the edge type set $\mathcal{T}$ includes the context, speaker, and utterance mode dependencies between utterances. The edge set ${\cal E}\subset{\cal V}\times{\cal V}$ indicates that there is a dialog relationship between nodes. The specific way to construct the graph is as follows:

\textbf{Nodes:} Given a dialog, the number of nodes in the graph is positively related to the number of sentences in the dialog. This paper constructed a graph containing $3N$ nodes, where each statement contains three nodes $v_i^a, v_i^v, v_i^t$ in the graph, which represent $[h_i^t,\lambda_i],[h_i^a,\lambda_i],[h_i^v,\lambda_i]$ and correspond to three modes.

\textbf{Edges:} We assume that every utterance in the same dialogue is related to other utterances. Therefore, there will be a connection between any two nodes of the same modality in the same dialog. Furthermore, each node is also connected to nodes of the same utterance but from different modalities. For example, in the graph node $v_i^t$ will be connected to $v_i^a$ and $v_i^v$.

\textbf{Edge Weighting:} We imagine that if there is a higher similarity between two nodes, then the information interaction between them will also be more important, which indicates that the edge weight between them should be larger. To capture the similarity between node expressions, we adopt angular similarity as a way to measure the edge weight of two nodes.

There are two types of edges in the graph, namely: 1) edges connecting nodes of the same modality, and 2) edges connecting nodes of different modalities. To differentiate these two cases, we employ different edge weighting strategies. For nodes of the same modality, the weights of their edges are calculated as follows:
\begin{equation}
\mathcal{A}_{ij}=1-\frac{arccos(sim(v_i,v_j))}{\pi}
\end{equation}
where $v_i$ and $v_j$ denote the feature representations of the $i$-th and $j$-th nodes in the graph. For nodes in different modalities, their edge weights are calculated as follows:
\begin{equation}
	\mathcal{A}_{ij}=\omega(1-\frac{arccos(sim(v_i,v_j))}{\pi})
\end{equation}
where $\omega$ is a hyper parameter.

According to the above steps, we further use GCN to encode contextual features.

\begin{equation}
\tilde{\mathcal{P}}=(\mathcal D+\mathcal I)^{-1/2}({A}_{ij}+\mathcal I)(\mathcal D+\mathcal I)^{-1/2}
\end{equation}
where $\mathcal{P}$ is the Laplacian matrix, $\mathcal D$ is the degree matrix, and $\mathcal I$ is the identity matrix. The iterative process of different layers can be represented by multiple layers of graph convolutional network (GCN) as follows:

\begin{equation}
\mathcal{H}^{(l+1)}=\varphi((1-\kappa)\tilde{\mathcal{P}}\mathcal{H}^{(l)}+\kappa\mathcal{H}^{(0)})((1-\varrho^{(l)})\mathcal{I+}\varrho^{(l)}\mathcal{W}^{(l)})
\end{equation}
where $\kappa$ and $\varrho$ represent hyperparameters, $\varphi$ represents the activation function, and $\mathcal{W}^{(l)}$ is a learned weight matrix.

\subsection{GCL: Graph Contrastive Learning}

GCL aims to make full use of the complementary information of different modalities by using contrastive learning, which can enhance the model's sensitivity to emotional features and improve emotion recognition accuracy by shortening the distance between positive samples and expanding the space between negative samples. For samples from the same modality, construct positive and negative pairs. Positive pairs include representations of the same modality for the same sample, and teams of negative samples have representations of the same modality for different samples. Using the Softmax function, the similarity between positive and negative examples is mapped to a specific range to distinguish intra-class and inter-class samples better, thereby providing a more accurate loss function for the training of comparative learning. The intra-class comparison loss and inter-class comparison loss are calculated as follows:

\begin{equation}
	\begin{aligned}
		\mathcal{L}_{MMD} = \frac{1}{N_p^2} &\sum_{i,j} \mu(\rho_i^\mathcal{M}, \rho_j^\mathcal{M}) \\
		- \frac{2}{N_pN_n} &\sum_{i,j} \mu(\rho_i^\mathcal{M}, \eta_j^\mathcal{M}) \\
		+ \frac{1}{N_n^2} &\sum_{i,j} \mu(\eta_i^\mathcal{M}, \eta_j^\mathcal{M})
	\end{aligned}
\end{equation}
where $N_p$ denotes the number of positive samples, $N_n$ denotes the
number of negative samples, $\rho_i^\mathcal{M}$ is a positive sample in the same category, $\eta_j^\mathcal{M}$ is a negative sample in the same category, $\mu$ is the kernel function, the similarity between the two texts.

The three terms in the formula calculate the difference in distribution between positive samples, the difference in distribution between positive and negative samples, and the difference in distribution between negative samples. By minimizing these distribution differences, we can make the model pay more attention to the emotion's characteristics rather than the differences between datasets. This helps improve the generalization ability of emotion classifiers to perform well under different data distributions.

Comparable loss and other possible classification loss combinations, reaching the loss function, the calculation process is as follows:

\begin{equation}
\mathcal{L}_{GCL}=\mathcal{L}_{MMN}+\zeta\cdot\mathcal{L}_{classification}
\end{equation}
Where $\zeta$ is the missing weight factor.

\subsection{Multi-task Learning}

After graph comparison learning processing, we get a vector $\chi_i^\mathcal{M}$ after multi-modal fusion. The target category (label) for each emotion is denoted by $y_i\in\{-1,0,1\}$, where $(-1)$, $(1)$, and $(0)$ represent the neutral emotion, positive emotion, and negative emotion of the first classification in the secondary classification, respectively.

\begin{table*}[!t]
	\renewcommand\arraystretch{1.5}
	\setlength{\tabcolsep}{14.8pt}
	\caption{On the IEMOCAP dataset, our method is compared with other baseline methods; the experimental results are shown in the table below. The best results in each column are shown in bold. Average(w) stands for weighted average.}
	\begin{tabular}{l|ccccccc}
		\hline
		\multirow{3}{*}{Methods} & \multicolumn{7}{c}{IEMOCAP}  \\ \cline{2-8}
		& Happy      & Sadness        & Neutral    & Angry      & Excitement    & Frustration & Average(w) \\ \cline{2-8}
		& Acc.  F1   & Acc.  F1   & Acc.  F1   & Acc.  F1   & Acc.  F1   & Acc.  F1   & Acc.  F1   \\ \hline
		bc-LSTM                  & 28.7  34.8 & 57.7  60.3 & 54.7  52.4 & 56.6  57.3 & 51.7  57.3 & 67.7  59.3 & 55.1  55.4 \\
		CMN                      & 25.7  30.1 & 55.6  62.8 & 53.2  52.7 & 61.0  59.3 & 55.0  60.3 & \textbf{70.7}  60.1 & 56.3  56.1 \\
		ICON                     & 22.5  30.3 & 59.1  64.9 & 62.6  57.8 & 65.1  63.4 & 58.8  62.9 & 67.1  60.7 & 59.7  59.0 \\
		MFN                      & 23.7  34.6 & 65.5  70.1 & 55.2  52.1 & 71.4  66.5 & 63.8  62.1 & 68.6  62.7 & 60.0  59.7 \\
		DialogueRNN              & 25.1  33.9 & 74.8  78.3 & 58.2  59.0 & 65.1  65.6 & 80.3  71.5 & 61.0  58.8 & 63.6  62.4 \\
		A-DMN                   & 43.0  50.2 & 69.8  76.4 & 63.0  62.9 & 63.6  56.5 & \textbf{87.8}  77.4 & 53.7  55.5 & 64.9  64.2 \\
		DialogueGCN              & 40.7  42.9 & \textbf{88.8  84.1} & 62.2  63.8 & 67.2  64.0 & 65.3  63.0 & 64.2  66.9 & 65.0  64.2 \\
		CTnet                   & 48.0  50.8 & 78.0  79.7 & \textbf{69.3  65.5} & \textbf{73.0} 67.2 & 85.6  \textbf{78.7} & 52.1  58.7 & 68.2  67.5 \\
		LR-GCN                  & 54.1  55.3 & 81.9  78.8 & 59.1  63.8 & 69.7  69.0 & 76.0  73.9 & 68.3  \textbf{68.6} & 68.5  68.1 \\
		GraphCFC                  & 43.5  54.1 & 85.1  84.5 & 64.3  62.0 & 71.2 \textbf{70.3} & 78.7  73.8 & 63.7  62.2 & 68.9  68.4 \\
		TS-GCL                   & \textbf{71.2  70.0} & 81.3  81.7 & {67.4 64.2}   & 60.5  61.4 & 74.6  76.5 & 62.0  64.6 & \textbf{70.3  70.2} \\ \hline
	\end{tabular}
	
	\label{tab2}
\end{table*}

The multi-modal fusion feature vector $\chi_i^\mathcal{M}$ and the corresponding target category $y_i$ will be used as the input of the multi-layer perceptron. By performing backpropagation training on samples, the multilayer perceptron will gradually learn appropriate weights and biases for better prediction in secondary classification tasks. The process is as shown in the formula:

\begin{equation}
y_i=softmax({MLP}(\chi_i^\mathcal{M})), y_i\in\{-1,0,1\}
\end{equation}

\begin{equation}
	\hat{y}_i=softmax({MLP}(\chi_i^\mathcal{M}))
\end{equation}

So far, we got emotion labels $\hat{y}_i$ for each sentence.

\section{Experiments}
\subsection{Implementation Details}

In this section, we describe the implementation details of the model during training. Our experimental environment is the Windows 11 operating system, and the computer used is equipped with an Intel Core i7 13700k processor and an NVIDIA RTX 3090 graphics card. The construction of the deep learning algorithm adopts Python 3.8 and PyTorch 1.9.1 version.

\subsection{Benchmark Dataset Used}

In MER, two multimodal dialogue data sets, IEMOCAP and MELD, are usually used for comparative experiments. The following is an introduction to these two data sets:

\textbf{IEMOCAP:} is an emotion database for studying emotional expressions and interactive behaviors. Dialogue in IEMOCAP covers a variety of emotional states, such as anger, happiness, sadness, neutral, etc., as well as different situations, such as face-to-face communication, telephone communication, etc. This allows researchers to explore the relationship between emotion, interaction, communication, etc. while examining multimodal expressions in a laboratory setting.

\textbf{MELD:} is a multimodal dataset widely used in emotion recognition research to help researchers understand emotional expressions more comprehensively. The dialogues in the dataset come from movie clips containing text dialogue, audio, and video information. The conversations of the MELD dataset cover a variety of emotional states, such as anger, happiness, sadness, neutral, etc., as well as different situations and emotional intensities.
\subsection{Baseline Models}

In this section, we detail the baseline model compared with the model in this paper, which are the results achieved on two general datasets, which will be described in detail below:

\textbf{text-CNN}  efficiently extracts critical text features through convolution pooling operation, and its end-to-end learning method also promotes the development of text classification tasks.

\textbf{MFN}  can give full play to multi-modal complementarity through multi-level fusion and end-to-end training, but its structure is complex, and training is difficult. A large amount of labeled data is required for supervised training.

\textbf{bc-LSTM:} The bc-LSTM provides sufficient context information through the target word's forward and backward context vectors to help the model better understand the semantics of the target word, but the computational complexity is also high.

\begin{table*}[!t]
	\centering
	\renewcommand\arraystretch{1.5}
	\caption{On the MELD dataset, our method is compared with other baseline methods. The experimental results are shown in the table below. The best results in each column are shown in bold. Average(w) stands for weighted average.}
	\setlength{\tabcolsep}{12.2pt}{
		\begin{tabular}{l|*{8}{c}}
			\hline
			\multirow{2}{*}{Methods} & \multicolumn{8}{c}{MELD} \\ \cline{2-9}
			& Neutral & Surprise & Fear & Sadness & Joy & Disgust & Anger & Average(w) \\
			& Acc. F1 & Acc. F1 & Acc. F1 & Acc. F1 & Acc. F1 & Acc. F1 & Acc. F1 & Acc. F1\\ \hline
			text-CNN & 74.8 73.2 & 45.2 45.1 & 3.7 3.2 & 21.7 22.1 & 49.8 48.7 & \textbf{8.7 8.6} & 35.0 34.7 & 55.3 54.9 \\
			MFN & 76.3 76.1 & 41.0 39.6 & 0.0 0.0 & 14.0 13.7 & 46.6 45.9 & 0.0 0.0 & 41.0 39.7 & 55.1 54.3 \\
			bc-LSTM & 76.8 76.1 & 25.0 24.5 & 9.2 8.6 & 24.2 23.6 & 54.4 54.1 & 4.4 4.1 & 44.0 42.9 & 59.5 58.3 \\
			CMN & 75.0 73.6 & 46.9 45.7 & 0.0 0.0 & 23.3 22.5 & 44.7 44.4 & 0.0 0.0 & 44.9 43.8 & 55.8 54.3 \\
			ICON & 74.2 73.6 & 50.1 48.8 & 0.0 0.0 & 23.3 22.4 & 50.4 49.9 & 0.0 0.0 & 45.1 43.8 & 56.6 55.2 \\
			DialogueRNN & 77.5 78.0 & 53.0 52.1 & 2.6 2.4 & 34.3 33.7 & 54.7 53.1 & 7.8 6.8 & 44.0 42.8 & 60.3 59.4 \\
			A-DMN & \textbf{78.7} 77.6 & 55.3 54.8 & 8.8 6.9 & 24.5 23.8 & 24.5 22.6 & 3.6 3.1 & 40.9 40.4 & 55.4 55.1 \\
			GraphCFC & 76.7 76.0 & 49.8 48.7 & 0.0 0.0 & 27.0 25.8 & 52.0 52.3 & 0.0 0.0 & \textbf{47.8 47.3} & 61.6 61.2 \\
			TS-GCL & 78.1 \textbf{80.6} & \textbf{56.7 56.4} & \textbf{6.8 5.2} & \textbf{42.3 43.7} & \textbf{68.3 66.3} & 2.3 2.6 & 43.8 48.5 & \textbf{64.4 64.1} \\ \hline
	\end{tabular}}
	\label{tab3}
\end{table*}
\textbf{CMN} proposed a cross-modal contextual attention mechanism, which can learn the correlation between text and video features and perform adaptive multi-modal fusion. However, it cannot clearly distinguish the feature contribution of unimodality and multimodality, and the demand for labeled data is significant.

\textbf{DialogueRNN} used a Conditional Random Field (CRF) in the context generator part, which can effectively model the context dependency of sentences. However, using CRF also increases the computational complexity of the model.

\textbf{DialogueGCN}  is a dialogue modeling method based on a graph convolutional network. It models dialogue as a sentence graph, and graph edges represent sentence relationships. GCN is used to learn the representation of sentence graphs. Therefore, it has achieved excellent performance on emotion recognition tasks.

\textbf{ICON} (Interaction-Context Network) is a network model for multimodal dialogue, which uses GRUs to track contextual interaction history so a better context receptive field is obtained.

\subsection{Results and Discussion}

We comprehensively compare our proposed emotion recognition algorithm TS-GCL with other deep learning algorithms. On the IEMOCAP and MELD datasets, Table I and Table II show the recognition accuracy and F1 value of each algorithm on each emotion category and the average accuracy and F1 value of the model as a whole. Experimental results significantly demonstrate the superior performance of our proposed algorithm.

\textbf{IEMOCAP:} As shown in Table I, TS-GCL has achieved excellent performance on the IEMOCAP dataset, taking the lead in four indicators, among which the accuracy rate and F1 are 70.3\% and 70.2\%, respectively. In addition, TS-GCL has achieved excellent performance in the happy category, and the accuracy and F1 of other categories are slightly lower than different existing algorithms. Analysis of the reasons shows that TS-GCL considers the similarities and differences between modalities and modalities during fusion and simulates human emotions for emotional classification by using graph comparison learning strategies and two emotions label classification methods for better performance.

\textbf{MELD:} As shown in Table II, TS-GCL has achieved excellent performance on the MELD dataset, leading in terms of indicators, among which the accuracy rate and F1 are 64.4\% and 64.1\%, respectively, compared with the existing comparison algorithms have achieved a small margin leading. In the four categories of surprise, fear, sadness, and Sadness, TS-GCL has achieved leading performance in accuracy and F1. Due to the severe category imbalance problem in the MELD dataset, TS-GCL generally performs on fear and disgust compared to existing algorithms, and our future work will also optimize this phenomenon.

The analysis of the aforementioned experimental results indicates that TS-GCL demonstrates superior performance, effectively capturing the similarities and differences among emotion samples. Based on these outcomes, further optimization can be pursued, along with the utilization of a secondary emotion classification approach. This approach avoids the shortcomings observed in previous methods for emotion classification.

\begin{figure}[htbp]
	\centering
	\includegraphics[width=0.94\linewidth,scale=1.0]{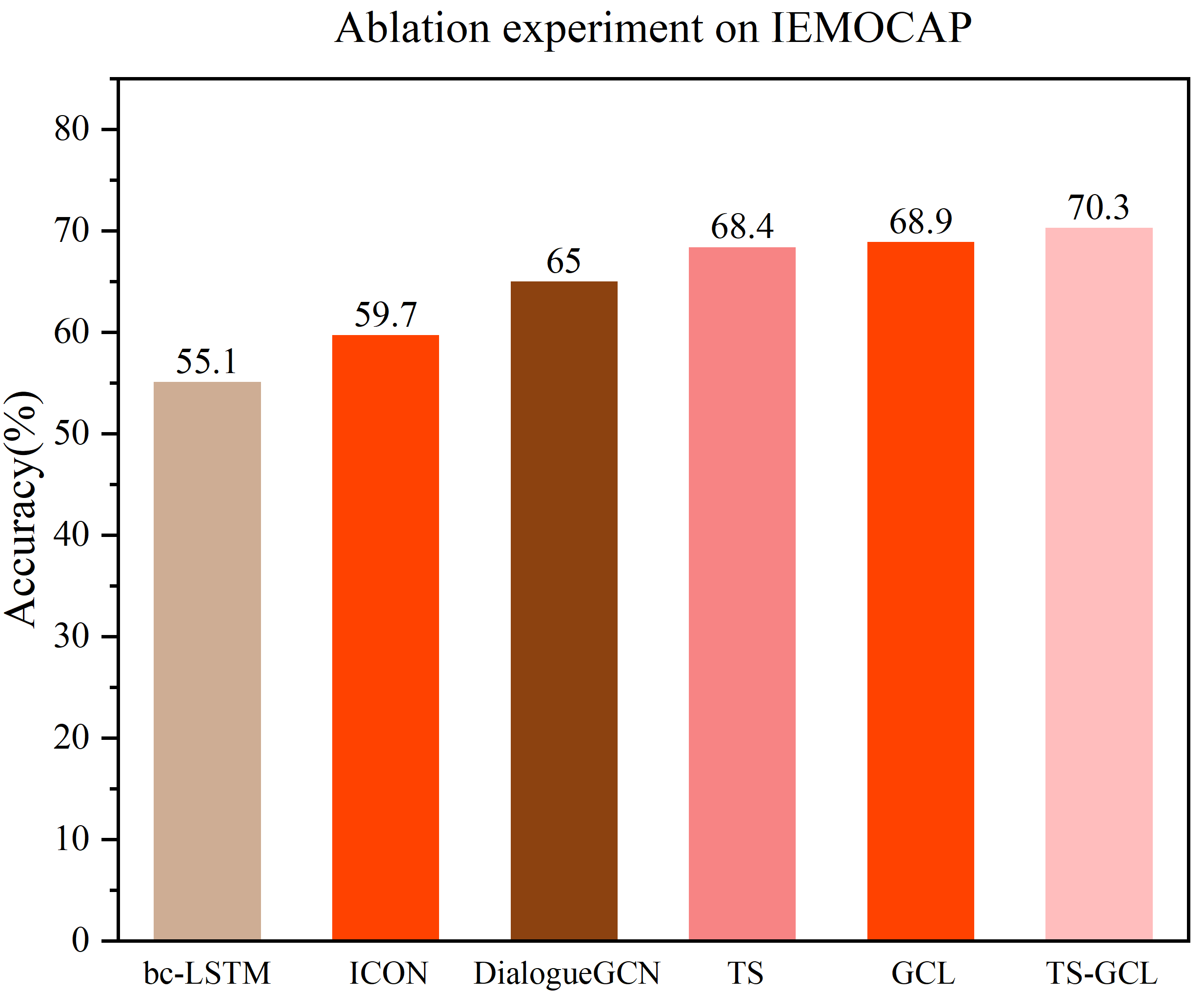}
	\caption{Ablation experiments on the IEMOCAP dataset. We conduct experiments on each component of TS-GCL and jointly compare with the benchmark models bc-LSTM, ICON and DialogueGCN.}
	\label{fig_1}
\end{figure}

\subsection{Ablation Studies}

In this section, we choose three benchmark models, dissect the TS-GCL model step by step, present the experimental results of each part, and analyze their performance on the IEMOCAP dataset to gain insight into their impact on the overall performance. At the same time, three benchmark models are selected for comparison. We present the corresponding experimental results in Fig. 3.

\begin{itemize}
	\item[$\bullet$] {Observing and comparing the effects of GCL and TS-GCL, we can see the impact of the secondary classification method on TS-GCL. Although related to our choice of state-of-the-art feature extractors, secondary classification methods still have some effect. Compared with not using TS (Two-Stage GCN), the accuracy is improved by 1.4\%. Secondary classification methods enable models to better adapt to specific domains or user emotions.}
	
	\item[$\bullet$] Observing and comparing the effects of TS and TS-GCL, we can see the influence of GCL on TS-GCL. Compared with not using GCL, the accuracy increased by 1.9\%. Graph contrastive learning introduces richer contextual information and relationship modeling for multimodal emotion recognition, which can improve the feature learning ability of the model, thereby achieving better results in multimodal emotion recognition tasks.
	
\end{itemize}

TS-GCL achieves the best performance when we use both TS and GCL. These partial combinations constitute our final model, and our plausible model is demonstrated experimentally.

\section{Conclusion }

In this paper, we propose a two-stage multimodal emotion recognition model based on graph contrastive learning (TS-GCL), which achieves efficient cross-modal feature fusion and designs a novel contrastive learning strategy to reduce the distance between samples. Different emotion labels in modals enhance the semantic representation ability of nodes. Then, we propose a two-stage classification method for multi-task multi-modal emotion recognition. The two-stage classification method can facilitate the module design of the model, so that modules in different stages can be tuned and optimized independently.

\section*{Acknowledgment}

This work is supported by National Natural
Science Foundation of China (Grant No. 61802444); the Changsha
Natural Science Foundation (Grant No. kq2202294), the Research
Foundation of Education Bureau of Hunan Province of China (Grant
No. 20B625, No. 18B196, No. 22B0275); the Research on Local
Community Structure Detection Algorithms in Complex Networks
(Grant No. 2020YJ009).


\begin{thebibliography}{00}



\bibitem{19} S. Qian, D. Xue, Q. Fang, and C. Xu, “Integrating multi-label contrastive learning with dual adversarial graph neural networks for cross-modal retrieval,” IEEE Transactions on Pattern Analysis and Machine Intelligence, pp. 1–18, 2022.


\bibitem{1} Shou, Y., Meng, T., Ai, W., Yang, S. and Li, K., 2022. Conversational emotion recognition studies based on graph convolutional neural networks and a dependent syntactic analysis. Neurocomputing, 501, pp.629-639.

\bibitem{2} Shou, Y., Cao, X., Meng, D., Dong, B. and Zheng, Q., 2023. A Low-rank Matching Attention based Cross-modal Feature Fusion Method for Conversational Emotion Recognition. arXiv preprint arXiv:2306.17799.
\bibitem{4} Y. Liu, M. Ott, N. Goyal, J. Du, M. Joshi, D. Chen, O. Levy, M. Lewis, L. Zettlemoyer, and V. Stoyanov, “Roberta: A robustly optimized bert pretraining approach,” arXiv preprint arXiv:1907.11692, 2019.
\bibitem{5} N. Majumder, S. Poria, D. Hazarika, R. Mihalcea, A. Gelbukh, and E. Cambria, “Dialoguernn: An attentive rnn for emotion detection in conversations,” in Proceedings of the AAAI conference on artificial intelligence, vol. 33, no. 01, 2019, pp. 6818–6825.
\bibitem{6} Z. Li, F. Tang, M. Zhao, and Y. Zhu, “Emocaps: Emotion capsule based model for conversational emotion recognition,” in Findings of the Association for Computational Linguistics: ACL 2022, 2022, pp.1610–1618.


\bibitem{17} F. Huang, X. Li, C. Yuan, S. Zhang, J. Zhang, and S. Qiao, “Attention-emotion-enhanced convolutional lstm for sentiment analysis,” IEEE Transactions on Neural Networks and Learning Systems, vol. 33, no. 9,pp. 4332–4345, 2022.

\bibitem{22} A. Zadeh, M. Chen, S. Poria, E. Cambria, and L.-P. Morency, “Tensor fusion network for multimodal sentiment analysis,” in Proceedings of the 2017 Conference on Empirical Methods in Natural Language Processing, 2017, pp. 1103–1114.

\bibitem{24} J. Hu, Y. Liu, J. Zhao, and Q. Jin, “Mmgcn: Multimodal fusion via deep graph convolution network for emotion recognition in conversation,”in Proceedings of the 59th Annual Meeting of the Association for Computational Linguistics and the 11th International Joint Conference on Natural Language Processing (Volume 1: Long Papers), 2021, pp.5666–5675.

\bibitem{12} S. Poria, E. Cambria, D. Hazarika, N. Majumder, A. Zadeh, and L.P. Morency, “Context-dependent sentiment analysis in user-generated videos,” in Proceedings of the 55th Annual Meeting of the Association for Computational Linguistics, vol. 1, 2017, pp. 873–883.
\bibitem{13} D. Hazarika, S. Poria, A. Zadeh, E. Cambria, L.-P. Morency, and R. Zimmermann, “Conversational memory network for emotion recognition in dyadic dialogue videos,” in Proceedings of the Conference of the North American Chapter ofthe Association for Computational Linguistics: Human Language Technologies, 2018, pp. 2122–2132.
\bibitem{14} M. Li, B. Yang, J. Levy, A. Stolcke, V. Rozgic, S. Matsoukas, C. Papayiannis, D. Bone, and C. Wang, “Contrastive unsupervised learning for speech emotion recognition,” in ICASSP 2021-2021 IEEE International Conference on Acoustics, Speech and Signal Processing (ICASSP).IEEE, 2021, pp. 6329–6333.
\bibitem{15} D. Kim and B. C. Song, “Contrastive adversarial learning for person independent facial emotion recognition,” in Proceedings of the AAAI Conference on Artificial Intelligence, vol. 35, no. 7. AAAI, 2021, pp.5948–5956.


\bibitem{30} Chen, C., Li, K., Li, Y. and Zou, X., 2022, April. ReGNN: A redundancy-eliminated graph neural networks accelerator. In 2022 IEEE International Symposium on High-Performance Computer Architecture (HPCA) (pp. 429-443). IEEE.
\bibitem{31} Chen, C., Li, K., Zou, X. and Li, Y., 2021, December. Dygnn: Algorithm and architecture support of dynamic pruning for graph neural networks. In 2021 58th ACM/IEEE Design Automation Conference (DAC) (pp. 1201-1206). IEEE.
\bibitem{32} Chen, C., Li, K., Teo, S.G., Zou, X., Li, K. and Zeng, Z., 2020. Citywide traffic flow prediction based on multiple gated spatio-temporal convolutional neural networks. ACM Transactions on Knowledge Discovery from Data (TKDD), 14(4), pp.1-23.

\bibitem{33} Meng, T., Shou, Y., Ai, W., Du, J., Liu, H., and Li, K., 2023. A multi-message passing framework based on heterogeneous graphs in conversational emotion recognition. Neurocomputing, 127109.

\bibitem{34} Shou, Y., Meng, T., Ai, W., Xie, C., Liu, H., and Wang, Y., 2022. Object Detection in Medical Images Based on Hierarchical Transformer and Mask Mechanism. Computational Intelligence and Neuroscience, 2022.

\bibitem{35} Shou, Y., Meng, T., Ai, W., Yin, N., and Li, K., 2023. A comprehensive survey on multi-modal conversational emotion recognition with deep learning. arXiv preprint arXiv:2312.05735.

\bibitem{36} Ying, R., Shou, Y., and Liu, C., 2021. Prediction Model of Dow Jones Index Based on LSTM-Adaboost. In 2021 International Conference on Communications, Information System and Computer Engineering (CISCE) (pp. 808-812). IEEE.

\bibitem{37} Shou, Y., Ai, W., and Meng, T., 2023. Graph information bottleneck for remote sensing segmentation. arXiv preprint arXiv:2312.02545.

\bibitem{38} Shou Y, Ai W, Meng T, et al. CZL-CIAE: CLIP-driven Zero-shot Learning for Correcting Inverse Age Estimation. arXiv preprint arXiv:2312.01758, 2023.

\bibitem{39} Meng T, Shou Y, Ai W, et al. Deep imbalanced learning for multimodal emotion recognition in conversations. arXiv preprint arXiv:2312.06337, 2023.

\bibitem{40} Ai, W., Shou, Y., Meng, T., and Li, K. 2023., DER-GCN: Dialogue and Event Relation-Aware Graph Convolutional Neural Network for Multimodal Dialogue Emotion Recognition. arXiv preprint arXiv:2312.10579.

\bibitem{41} Shou, Y., Meng, T., Ai, W., and Li, K. 2023., Adversarial Representation with Intra-Modal and Inter-Modal Graph Contrastive Learning for Multimodal Emotion Recognition. arXiv preprint arXiv:2312.16778.





\end{thebibliography}
\end{document}